\documentclass[letterpaper]{article} 
\usepackage{aaai2026}  
\usepackage{times}  
\usepackage{helvet}  
\usepackage{courier}  
\usepackage[hyphens]{url}  
\usepackage{graphicx} 
\usepackage{natbib}  
\usepackage{caption} 
\usepackage{algorithm}
\usepackage{amsfonts} 
\usepackage{amsmath}
\usepackage{algpseudocode}
\usepackage{multirow}
\usepackage{newfloat}
\usepackage{listings}
\DeclareCaptionStyle{ruled}{labelfont=normalfont,labelsep=colon,strut=off} 
\floatstyle{ruled}
\newfloat{listing}{tb}{lst}{}
\floatname{listing}{Listing}
\title{Boosting the Robustness-Accuracy Trade-off of SNNs \\ by Robust Temporal Self-Ensemble}
\author{
    Jihang Wang\textsuperscript{\rm 1,\rm 2},
    Dongcheng Zhao\textsuperscript{\rm 1,\rm 4,\rm 5},
    Ruolin Chen\textsuperscript{\rm 1,\rm 2},
    Qian Zhang\thanks{Corresponding author} \textsuperscript{\rm 1,\rm 2,\rm 4,\rm 5},
    Yi Zeng\thanks{Corresponding author} \textsuperscript{\rm 1,\rm 2,\rm 3,\rm 4,\rm 5}
}
\affiliations{
    \textsuperscript{\rm 1}Brain-inspired Cognitive AI Lab, Institute of Automation, Chinese Academy of Sciences, Beijing, China. \\
    \textsuperscript{\rm 2}School of Artificial Intelligence, University of Chinese Academy of Sciences, Beijing, China. \\
    \textsuperscript{\rm 3}School of Future Technology, University of Chinese Academy of Sciences, Beijing, China. \\
    \textsuperscript{\rm 4} Beijing Key Laboratory of Safe AI and Superalignment, Beijing, China. \\
    \textsuperscript{\rm 5} Center for Long-term AI, Beijing, China. \\
    yi.zeng@ia.ac.cn
}

\begin{document}

\maketitle

\begin{abstract}
Spiking Neural Networks (SNNs) offer a promising direction for energy-efficient and brain-inspired computing, yet their vulnerability to adversarial perturbations remains poorly understood. In this work, we revisit the adversarial robustness of SNNs through the lens of temporal ensembling, treating the network as a collection of evolving sub-networks across discrete timesteps. This formulation uncovers two critical but underexplored challenges—the fragility of individual temporal sub-networks and the tendency for adversarial vulnerabilities to transfer across time. To overcome these limitations, we propose Robust Temporal self-Ensemble (RTE), a training framework that improves the robustness of each sub-network while reducing the temporal transferability of adversarial perturbations. RTE integrates both objectives into a unified loss and employs a stochastic sampling strategy for efficient optimization. Extensive experiments across multiple benchmarks demonstrate that RTE consistently outperforms existing training methods in robust-accuracy trade-off. Additional analyses reveal that RTE reshapes the internal robustness landscape of SNNs, leading to more resilient and temporally diversified decision boundaries. Our study highlights the importance of temporal structure in adversarial learning and offers a principled foundation for building robust spiking models.
\end{abstract}


\section{Introduction}

Deep Neural Networks (DNNs) have achieved remarkable success in processing high-dimensional data and delivering state-of-the-art performance across a wide range of tasks~\cite{goodfellow2016deep,he2016deep}. These models typically rely on continuous-valued activations for inter-layer communication. While effective, this design departs significantly from the signaling mechanisms of biological neural systems and incurs high energy costs, particularly in large-scale or resource-constrained settings. 

To address this gap, Spiking Neural Networks (SNNs) have emerged as a brain-inspired computational paradigm that more faithfully captures the temporal dynamics and event-driven nature of biological neurons~\cite{maass1997networks,shen2023brain}. In contrast to conventional DNNs, SNNs transmit information via discrete spike trains distributed across space and time. Their asynchronous, event-triggered processing and multiplication-free synaptic operations enable substantially higher energy efficiency, particularly when deployed on neuromorphic hardware designed to exploit their sparse, binary signaling patterns~\cite{pei2019towards,roy2019towards,li2023firefly}.

Despite these advantages, SNNs have been shown to be similarly vulnerable to adversarial perturbations as their DNN counterparts~\cite{bu2023rate,hao2023threaten,liang2021exploring,lun2025towards}. This susceptibility highlights a critical gap between current SNN models and the robustness of biological neural systems, limiting their reliability in safety-critical applications. To improve adversarial robustness of SNNs, most existing work has focused on adversarial training (AT) with small perturbation budgets~\cite{ding2022snn,ding2024enhancing,ding2024robust,liu2024enhancing}, typically $\varepsilon = 2/255$ or $4/255$ on standard benchmarks such as CIFAR-10 and CIFAR-100~\cite{krizhevsky2009learning}. Although AT has been widely adopted in both artificial and spiking neural networks~\cite{madry2017towards}, small-budget perturbations often fail to sufficiently explore the neighborhood of training examples, resulting in limited robustness.
\begin{figure}[t]
  \centering
  \includegraphics[width=0.95\columnwidth]{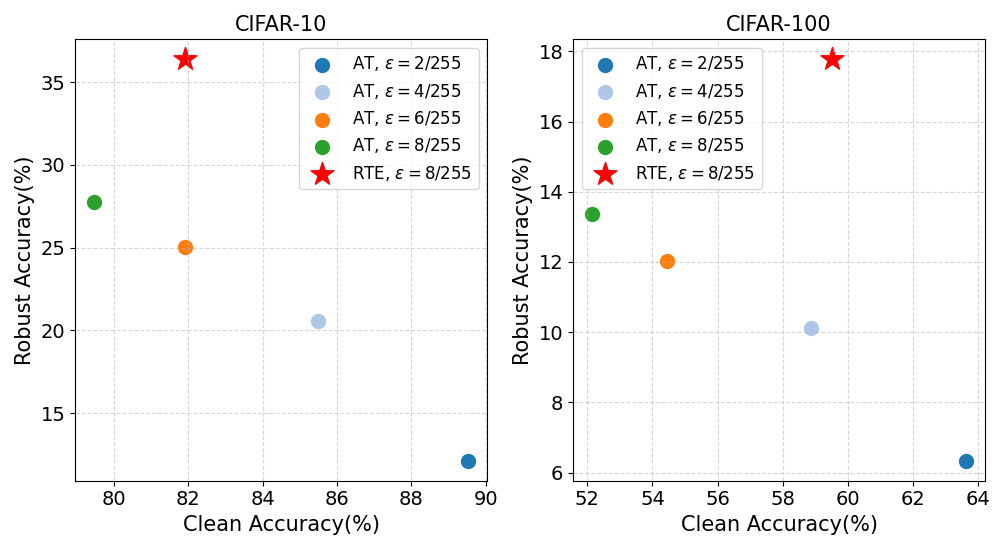}
  \caption{Comparison of clean and robust accuracy under AutoPGD attacks for AT and RTE on CIFAR-10 and CIFAR-100.}
  \label{fig1}
\end{figure}

A straightforward remedy is to increase the perturbation budget $\varepsilon$ during training. Empirical results show that this can improve robust accuracy, but often at the expense of a substantial drop in clean accuracy, as shown in Figure~\ref{fig1}. This trade-off becomes especially severe on more challenging datasets such as CIFAR-100, where stronger perturbations degrade generalization on clean inputs despite robustness gains. These observations reveal a fundamental limitation of current AT strategies for SNNs—simply increasing attack strength is insufficient to achieve a favorable robustness–accuracy balance.

To improve the robustness–accuracy trade-off, we present a novel training framework grounded in a temporal ensemble interpretation of SNNs. Specifically, we formalize the output of a spiking network as an implicit ensemble of temporal sub-networks, each corresponding to neural activity at a specific timestep~\cite{ding2025rethinking,zhao2025improving}. This formulation reveals a natural inductive structure within SNNs, wherein predictions emerge from a composition of temporally distributed decision paths. Leveraging this insight, we introduce \textbf{Robust Temporal self-Ensemble (RTE)}, a training methodology that jointly (i) improves the adversarial robustness of individual temporal sub-networks and (ii) suppresses the transferability of vulnerabilities across timesteps. RTE achieves this via an adjustable regularization scheme that operates across temporal components, effectively mitigating temporal vulnerabilities to adversarial directions.  Our main contributions are summarized as follows:
\begin{itemize}
\item We formulate the output of an SNN as a temporal self-ensemble, and propose the RTE framework to simultaneously improve the robustness of individual sub-networks and suppress vulnerability transfer across timesteps.

\item RTE introduces an adjustable regularization mechanism that enables SNNs to explore broader local neighborhoods in the input space while maintaining high clean accuracy with minimal compromise.

\item Extensive evaluations on standard benchmarks demonstrate that RTE is a scalable and effective defense strategy. It consistently achieves a better robustness–accuracy trade-off, particularly on more complex datasets like CIFAR-100.
\end{itemize}

\section{Related Work}

\subsection{Adversarial Defense in Spiking Neural Networks}
Efforts to improve the adversarial robustness of SNNs fall into three broad categories: regularization-based training, stochastic modeling, and input encoding strategies. Regularization-based methods introduce additional loss terms or impose architectural constraints to stabilize model behavior under perturbations. For instance, Regularized Adversarial Training (RAT) applies parseval regularization to bound the Lipschitz constant~\cite{ding2022snn}, Membrane Potential Perturbation Dynamics (MPPD) leverages neuron dynamics to stabilize SNNs~\cite{ding2024robust}, and Sparsity Regularization (SR) uses sparsity-aware smoothing techniques~\cite{liu2024enhancing}. However, such techniques often rely on handcrafted priors and lack temporal specificity. Stochastic modeling injects structured randomness into network dynamics to improve robustness. Examples include randomized smoothing for rate-coded SNNs~\cite{mukhotycertified}, stochastic synaptic gating~\cite{ding2024enhancing}, and heterogeneity in time constants~\cite{wang2025random}. While effective in certain settings, these methods may introduce significant inference variance and do not explicitly mitigate inter-timestep vulnerabilities. Encoding-based strategies seek to enhance robustness by modifying spike-based input representations. Techniques such as signed rate encoding~\cite{mukhoty2025improving} and frequency-domain transformations like FEEL-SNN~\cite{xu2024feel} aim to distribute or normalize input information. Nevertheless, they primarily address input-level perturbations and do not provide internal robustness guarantees.

\subsection{Ensemble-Based Robustness and Temporal Modeling in SNNs}

Ensemble-based adversarial defenses have shown promise in improving robustness by promoting prediction diversity and reducing cross-model vulnerability transfer~\cite{pang2019improving,yang2020dverge,yang2021trs,deng2023understanding}. For example, TRS maximizes gradient disagreement across ensemble members~\cite{yang2021trs}, and DVERGE explicitly minimizes overlap in adversarial subspaces~\cite{yang2020dverge}. Recently, several works have extended ensemble concepts to temporal modeling in SNNs, interpreting network outputs as implicit ensembles across timesteps~\cite{ding2025rethinking,zhao2025improving}. These methods typically focus on enforcing temporal consistency or stability. However, they often overlook inter-timestep diversity and fail to explicitly suppress temporal vulnerability transfer—an issue particularly critical under ensemble-based robustness. In contrast, our proposed RTE framework builds on a principled temporal ensemble view of SNNs. RTE directly targets the robustness of individual temporal sub-networks and introduces regularization to decouple adversarial influence across time. This formulation enables scalable and architecture-agnostic adversarial defense, achieving superior robustness–accuracy trade-offs compared to prior AT-based and ensemble-inspired methods.

\section{Notations and Preliminaries}

\subsection{Adversarial Attacks}

We consider untargeted adversarial attacks constrained by an $\ell_p$-norm ball. Let $\boldsymbol{f}:\mathbb{R}^d \rightarrow \mathbb{R}^C$ denote a multi-class classifier, and let $\boldsymbol{x} \in \mathbb{R}^d$ be an input sample with ground-truth label $y$. Given a perturbation budget $\varepsilon > 0$, an adversarial example $\boldsymbol{x'}$ is defined as the solution to the following optimization problem:
\begin{equation}
    \boldsymbol{x'} = \arg \max_{\left\lVert \boldsymbol{x'} - \boldsymbol{x} \right\rVert_p \leq \varepsilon} \mathcal{L}(\boldsymbol{f}(\boldsymbol{x'}), y)
\label{1}
\end{equation}
where $\mathcal{L}$ denotes the cross-entropy (CE) loss. Standard attack algorithms include FGSM~\cite{goodfellow2014explaining}, PGD~\cite{madry2017towards}, and AutoAttack~\cite{croce2020reliable}, the latter of which combines multiple strong attacks (e.g., AutoPGD) to provide reliable worst-case robustness estimates.

\subsection{Spiking Neural Networks}

Spiking Neural Networks process information over discrete timesteps $t = 1,\dots,T$ using binary spike trains. Each layer $l$ produces a spike output $\boldsymbol{s}^l(t) \in \{0,1\}$ governed by the Leaky Integrate-and-Fire (LIF) neuron model:
\begin{equation}
    \boldsymbol{V}^l(t+1) = \lambda \boldsymbol{V}^l(t) \odot (1 - \boldsymbol{s}^l(t)) + (1 - \lambda) \boldsymbol{W}^l \boldsymbol{s}^{l-1}(t+1)
\end{equation}
\begin{equation}
    \boldsymbol{s}^l(t+1) = H(\boldsymbol{V}^l(t+1) - V_{\text{th}})
\end{equation}
where $\boldsymbol{V}^l(t)$ denotes the membrane potential, $\lambda$ is the leak rate, $\boldsymbol{W}^l$ is the synaptic weight matrix, and $H(\cdot)$ is the Heaviside step function, acting as a binary thresholding mechanism.

For static inputs such as images, direct encoding is commonly employed, where the same input $\boldsymbol{x}$ is injected into the network at every timestep. The final prediction is computed by aggregating output activations over $T$ timesteps:
\begin{equation}
    \boldsymbol{f}(\boldsymbol{x}) = \frac{1}{T} \sum_{t=1}^{T} \boldsymbol{W}^L \boldsymbol{s}^{L-1}(t)
    \label{output}
\end{equation}

Due to the non-differentiability of $H(\cdot)$, training SNNs typically relies on surrogate gradient techniques, which provide differentiable approximations to $H(\cdot)$ and allow optimization via backpropagation through time. We provide the specific surrogate gradient functions in Appendix A.

\section{Methods}

The output of an SNN, as defined in Eq.~\ref{output}, can be reformulated as a temporal average over per-timestep predictions:
\begin{equation}
    \boldsymbol{f}(\boldsymbol{x}) = \frac{1}{T} \sum_{t = 1}^{T} \boldsymbol{f}_t(\boldsymbol{x}), \quad \text{where} \quad \boldsymbol{f}_t(\boldsymbol{x}) = \boldsymbol{W}^L \boldsymbol{s}^{L-1}(t)
\end{equation}
This formulation induces a natural interpretation of SNNs as temporal ensembles, where each $\boldsymbol{f}_t(\boldsymbol{x})$ corresponds to the network's prediction at timestep $t$. We define each $\boldsymbol{f}_t(\boldsymbol{x})$ as the output of a distinct \textit{temporal sub-network}.

Leveraging this temporal ensemble structure, we introduce an adversarial training framework that jointly optimizes two complementary objectives: (i) reducing the individual adversarial vulnerability of each temporal sub-network, and (ii) minimizing the transferability of adversarial perturbations across timesteps.

\begin{figure}[t] 
  \centering 
  \includegraphics[width=0.8\columnwidth]{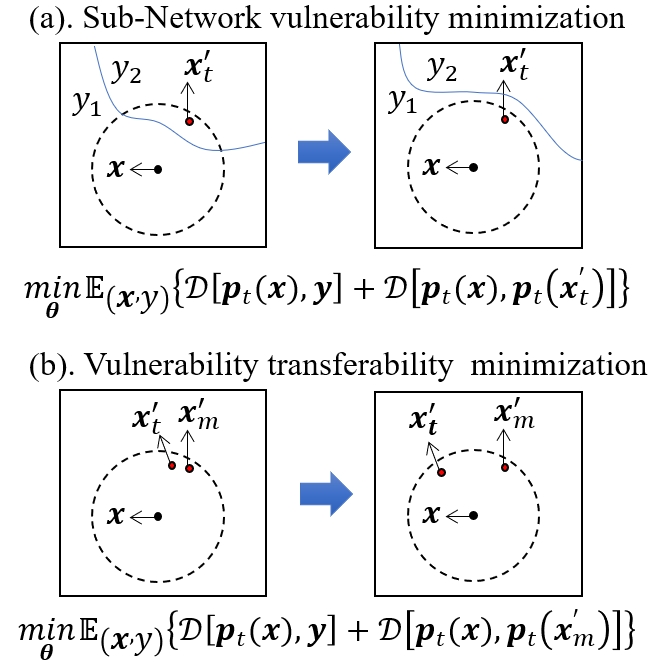}
\caption{
(a) Sub-network vulnerabilities are characterized by their most sensitive input perturbation $\boldsymbol{x}'_t$.
(b) Minimizing vulnerability transferability reduces shared weakness across sub-networks, enhancing ensemble robustness.
}
  \label{fig2} 
\end{figure}
\subsection{Sub-Network Vulnerability Minimization}

Most existing approaches evaluate adversarial robustness solely based on the aggregated SNN output $\boldsymbol{f}(\boldsymbol{x})$, without explicitly accounting for vulnerabilities at individual timesteps. If we adopt cross-entropy (CE) loss in Eq.~\ref{1}, we obtain the following upper bound:
\begin{align}
\max_{\left\lVert \boldsymbol{x'} - \boldsymbol{x} \right\rVert_p \leq \varepsilon} \mathcal{L}_\mathrm{CE}(\boldsymbol{f}(\boldsymbol{x'}), y) \notag \\
&\hspace{-8em} \leq \max_{\left\lVert \boldsymbol{x'} - \boldsymbol{x} \right\rVert_p \leq \varepsilon} \sum_{t=1}^T \frac{\mathcal{L}_\mathrm{CE}(\boldsymbol{f}_t(\boldsymbol{x'}), y)}{T} \notag \\
&\hspace{-8em} \leq \sum_{t=1}^T \max_{\left\lVert \boldsymbol{x'} - \boldsymbol{x} \right\rVert_p \leq \varepsilon} \frac{\mathcal{L}_\mathrm{CE}(\boldsymbol{f}_t(\boldsymbol{x'}), y)}{T}
\label{inq1}
\end{align}
The first inequality follows from~\cite{deng2022temporal}. Eq.~\ref{inq1} reveals that evaluating adversarial robustness based solely on $\boldsymbol{f}(\boldsymbol{x})$ may underestimate the vulnerabilities of individual temporal sub-networks $\boldsymbol{f}_t(\boldsymbol{x})$.

Let $\boldsymbol{p}_t(\boldsymbol{x}) = \sigma(\boldsymbol{f}_t(\boldsymbol{x}))$ denote the predicted class distribution at timestep $t$, where $\sigma$ is the softmax function. We quantify the maximum perturbation-induced distributional shift using a distance metric $\mathcal{D}$:
\begin{equation}
\max_{\left\lVert \boldsymbol{x}' - \boldsymbol{x} \right\rVert_p \leq \varepsilon} \mathcal{D}[\boldsymbol{p}_t(\boldsymbol{x}), \boldsymbol{p}_t(\boldsymbol{x}')]
\label{Ltt}
\end{equation}

Moreover, the adversarial loss of $\boldsymbol{f}_t(\boldsymbol{x})$ under $\mathcal{D}$ can be upper-bounded via the triangle inequality:
\begin{align}
\max_{\left\lVert \boldsymbol{x}' - \boldsymbol{x} \right\rVert_p \leq \varepsilon} \mathcal{D}[\boldsymbol{p}_t(\boldsymbol{x}'), \boldsymbol{y}] 
&\leq \mathcal{D}[\boldsymbol{p}_t(\boldsymbol{x}), \boldsymbol{y}] \notag \\
&\quad + \max_{\left\lVert \boldsymbol{x}' - \boldsymbol{x} \right\rVert_p \leq \varepsilon} \mathcal{D}[\boldsymbol{p}_t(\boldsymbol{x}), \boldsymbol{p}_t(\boldsymbol{x}')]
\label{inq2}
\end{align}

Thus, minimizing the right-hand side of Inequality~\ref{inq2} serves as an effective strategy to reduce the adversarial vulnerability of temporal sub-networks (Figure~\ref{fig2}a).

\subsection{Vulnerability Transferability Minimization}

To further mitigate vulnerability propagation across temporal sub-networks, we define $\boldsymbol{x}'_t$ as the input within an $\ell_p$-bounded neighborhood of $\boldsymbol{x}$ that induces the maximum distributional shift for sub-network $\boldsymbol{f}_t$:
\begin{equation}
\boldsymbol{x}'_t = \arg \max_{\left\lVert \boldsymbol{x}' - \boldsymbol{x} \right\rVert_p \leq \varepsilon} \mathcal{D}[\boldsymbol{p}_t(\boldsymbol{x}), \boldsymbol{p}_t(\boldsymbol{x}')]
\label{x't}
\end{equation}

We evaluate transferability by applying $\boldsymbol{x}'_t$ to another temporal sub-network $\boldsymbol{f}_m$ $(m \ne t)$. The resulting adversarial loss can be upper bounded via the triangle inequality:
\begin{align}
\mathcal{D}[\boldsymbol{p}_m(\boldsymbol{x}'_t), \boldsymbol{y}] 
&\leq \mathcal{D}[\boldsymbol{p}_m(\boldsymbol{x}), \boldsymbol{y}] 
+ \mathcal{D}[\boldsymbol{p}_m(\boldsymbol{x}), \boldsymbol{p}_m(\boldsymbol{x}'_t)]
\label{inq3}
\end{align}

We minimize the right-hand side of Eq.~\ref{inq3} to reduce the vulnerability transferability from $\boldsymbol{f}_t$ to $\boldsymbol{f}_m$. A lower value indicates that $\boldsymbol{f}_t$'s vulnerability does not significantly affect $\boldsymbol{f}_m$, implying disjoint adversarial regions and reduced temporal correlation (Figure~\ref{fig2}b).

By combining Eq.~\ref{inq2}-\ref{inq3}, we can quantify the cross-timestep distributional shift of the temporal sub-networks in SNNs using a transferability matrix:
\begin{equation}
\mathcal{L}^{(t,m)} = \mathcal{D}[\boldsymbol{p}_m(\boldsymbol{x}), \boldsymbol{p}_m(\boldsymbol{x}'_t)]
\label{ltm}
\end{equation}
where the diagonal elements of the matrix $\mathcal{L}^{(t,t)}$ represent the self-robustness of each sub-network $\boldsymbol{f}_t(\boldsymbol{x})$. Lower $\mathcal{L}^{(t,t)}$ values indicate stronger resistance to perturbations within the neighborhood of $\boldsymbol{x}$ (Figure~\ref{fig2}a). The off-diagonal elements $\mathcal{L}^{(t,m)}(t \neq m)$ represent the inter-subnetwork transferability of distributional shift. Smaller values reflect lower inter-timestep correlation and higher temporal diversity.



\subsection{Robust Temporal Self-Ensemble}

Building upon the vulnerability metrics in Eq.~\ref{inq2} and \ref{inq3}, we formulate an adversarial training objective that simultaneously mitigates the intrinsic vulnerability of each temporal sub-network and suppresses the transferability of adversarial perturbations across timesteps. Specifically, we define the Robust Temporal self-Ensemble (RTE) objective as:
\begin{equation}
\min_{\boldsymbol{\theta}} \mathbb{E}_{(\boldsymbol{x}, y)} \sum_{t=1}^{T} \sum_{m=1}^{T} \left\{ \mathcal{D}[\boldsymbol{p}_t(\boldsymbol{x}), \boldsymbol{y}] + \mathcal{D}[\boldsymbol{p}_t(\boldsymbol{x}), \boldsymbol{p}_t(\boldsymbol{x}'_m)] \right\}
\label{o1}
\end{equation}
Here, $\boldsymbol{\theta}$ denotes the trainable parameters of the SNN. The first term encourages accurate predictions at each timestep, while the second term jointly reduces sub-network vulnerability and inter-timestep vulnerability alignment.

Directly computing the adversarial examples $\boldsymbol{x}'_m$ for all $m = 1,\dots,T$ is computationally prohibitive, particularly due to the temporal coupling introduced by Batch Normalization (BN) layers, which require full-sequence inputs across $T$ steps for a single forward pass~\cite{duan2022temporal}.

To alleviate this overhead, we adopt a stochastic sampling strategy. At each training iteration, a single timestep $m$ is randomly selected to generate the adversarial example $\boldsymbol{x}'_m$. We then optimize the following surrogate objective:
\begin{equation}
\min_{\boldsymbol{\theta}} \mathbb{E}_{(\boldsymbol{x}, y)} \sum_{t=1}^{T} \left\{ \mathcal{D}[\boldsymbol{p}_t(\boldsymbol{x}), \boldsymbol{y}] + \mathcal{D}[\boldsymbol{p}_t(\boldsymbol{x}), \boldsymbol{p}_t(\boldsymbol{x}'_m)] \right\}
\label{o2}
\end{equation}
This approximation substantially reduces computational cost by limiting the perturbation generation to a single sub-network per iteration, while still offering effective gradient signals for improving robustness.

In practice, the raw distance metric $\mathcal{D}$ may be non-smooth and suboptimal for training. We therefore adopt a smoothed variant—its squared form—and further relax it using the Pinsker inequality when $\mathcal{D}$ is the $\ell_1$ distance:
\begin{equation}
\frac{1}{2}\|\boldsymbol{p} - \boldsymbol{q}\|_1^2 \leq \operatorname{KL}(\boldsymbol{p} \| \boldsymbol{q})
\label{pinsky}
\end{equation}
This yields a KL-based formulation that aligns with distillation-style regularization~\cite{ding2025rethinking}, enhancing both optimization stability and theoretical interpretability. We introduce a regularization coefficient $\gamma$ to balance clean and robust performance, and define the final training loss of RTE as:
\begin{equation}
\mathcal{L}_{\mathrm{RTE}} = \frac{1}{T} \sum_{t=1}^{T} \left\{ \operatorname{KL}[\boldsymbol{p}_t(\boldsymbol{x}) \| \boldsymbol{y}] + \gamma \operatorname{KL}[\boldsymbol{p}_t(\boldsymbol{x}) \| \boldsymbol{p}_t(\boldsymbol{x}'_m)] \right\}
\label{rte}
\end{equation}
The overall training procedure is summarized in Algorithm~\ref{alg1}.

\begin{algorithm}[!tb]
\caption{Robust Temporal self-Ensemble Training}\label{alg1}
\begin{algorithmic}[1]
\Require SNN with $T$ 
timesteps, trainable parameter $\boldsymbol{\theta }$, training epochs $E$, number of batched input-label pairs $B$, learning rate $\eta $, perturbation magnitudes $\varepsilon $, PGD iteration steps $K$, PGD step size $\alpha $.
\State Initialization SNN
\For{$e = 1$ \textbf{to} $E$} \Comment{Training iteration}
    \For{$b = 1$ \textbf{to} $B$}
        \State Sample batched input data $(\boldsymbol{x}, y) $
        \State Randomly sample $m \in \{1,2,...,T\}$
        \State $\boldsymbol{x'_m} \leftarrow \boldsymbol{x} + \boldsymbol{U}(-\varepsilon ,\varepsilon )$, where $\boldsymbol{U}$ denotes sampling from a uniform distribution.
        \For{$k = 1$ \textbf{to} $K$}
            \State Update $\boldsymbol{x'_m}$ by $\boldsymbol{x'_m} \leftarrow \Pi_{\mathbb{B}\left(\boldsymbol{x}, \varepsilon\right)}\left[\alpha \operatorname{sign}\left(\nabla_{\boldsymbol{x'_m}} \operatorname{KL}[\boldsymbol{p}_m(\boldsymbol{x})\| \boldsymbol{p}_m(\boldsymbol{x}'_m)]\right)\right]+\boldsymbol{x'_m}$, where $\Pi_{\mathbb{B}\left(\boldsymbol{x}, \varepsilon\right)}$ projects the input onto the bounded interval $\mathbb{B}\left(\boldsymbol{x}, \varepsilon\right)$.
        \EndFor
        \State Calculate $\mathcal{L}_{\mathrm{RTE}}$ using $\boldsymbol{x'_m}$ as Eq. \ref{rte} 
        \State Calculate gradients via backpropagation and update the parameters:\\ $$\boldsymbol{\theta } \leftarrow \boldsymbol{\theta } - \eta \nabla _{\boldsymbol{\theta } } \mathcal{L}_{\mathrm{RTE}}$$
    \EndFor
\EndFor
\end{algorithmic}
\end{algorithm}
\begin{table*}[t]
  \centering
  \begin{tabular}{c|c|ccccccc}
  \hline
 Datasets & Methods & Clean & FGSM & PGD$_{10}$ & PGD$_{50}$ & APGD$^{CE}_{50}$ & APGD$^{DLR}_{50}$ & Square \\
   \hline
 \multirow{7}{*}{\shortstack{SVHN \\ T=4 }} & AT & 92.81 & 67.90 & 54.42 & 49.79 & \underline{40.99} & 44.19 & 51.30 \\
    & RAT & 92.78 & 66.71 & 55.45 & 51.53 & \underline{44.59} & 44.69 & 56.71  \\
    & AT+MPPD & 92.62 & 66.78 & 53.51 & 49.79 & \underline{40.33} & 42.81 & 50.11  \\
    & AT+SR & 90.68 & 68.36 & \textbf{62.17} & \textbf{60.31} & \textbf{53.63} & \underline{48.16} & 53.98 \\
    & TRADES & \textbf{94.00} & \textbf{71.98} & 59.36 & 55.30 & 48.58 & \underline{48.14} & 57.04 \\
 \cline{2-9}
    & RTE& 93.59 & 71.38 & 59.78 & 56.56 & 50.63 & \underline{50.06} & 58.43  \\
    & RTE+SR & 92.82 & 69.43 & {59.89} & {57.24} & {52.36} & \underline{\textbf{51.03}} & \textbf{58.54}  \\
    \hline
\multirow{7}{*}{\shortstack{CIFAR \\ -10 \\ T=4}} & AT & 79.04 & 51.17 & 44.99 & 41.87 & \underline{26.92} & 29.75 & 32.66 \\
    & RAT & 80.24 & 52.70 & 46.81 & 44.41 & \underline{31.71} & 34.74 & 41.99 \\
    & AT+MPPD & 79.20 & 50.87 & 44.63 & 41.96 & \underline{26.59} & 30.03 & 31.43 \\
    & AT+SR & 78.55 & \textbf{55.01} & \textbf{51.61} & \textbf{50.82} & \textbf{41.40} & \underline{\textbf{41.21}} & 49.36 \\
    & TRADES & 81.80 & 53.03 & 47.09 & 45.37 & \underline{34.68} & 36.79 & 46.53 \\
 \cline{2-9}
& RTE & \textbf{81.90} & {53.14} & {48.07} & {46.33} & \underline{36.38} & {38.60} & {47.05} \\
    & RTE+SR & 80.85 & 53.45 & {49.05} & {47.73} & \underline{40.31} & {40.55} & \textbf{52.79} \\
  \hline
\multirow{7}{*}{\shortstack{CIFAR \\ -100 \\ T=4}} & AT & 51.82 & 25.86 & 22.20 & 21.07 & \underline{13.08} & 15.05 & 17.20 \\
    & RAT & 54.99 & \textbf{28.96} & 25.69 & 24.38 & \underline{17.34} & 18.65 & 23.98 \\
    & AT+MPPD & 52.09 & 25.07 & 21.77 & 20.81 & \underline{13.05} & 15.20 & 18.11 \\
    & AT+SR & 51.72 & 28.44 & \textbf{26.12} & 25.07 & \underline{18.57} & 19.23 & 23.58 \\
    & TRADES & 56.97 & 27.67 & 24.70 & 23.95 & 17.66 & \underline{17.39} & 24.69 \\
 \cline{2-9}
    & RTE & \textbf{59.50} & {28.66} & {25.45} & {24.28} & \underline{17.77} & {18.80} & {26.60} \\
    & RTE+SR & 58.10 & 28.90 & {26.07} & \textbf{25.52} & \textbf{20.18} & \underline{\textbf{19.95}} & \textbf{30.24} \\

       \hline
\multirow{7}{*}{\shortstack{Tiny- \\ ImageNet \\ T=4}} & AT & 42.36 & 24.12 & 22.50 & 22.17 & \underline{10.93} & 12.58 & 12.47 \\
    & RAT & 46.58 & \textbf{29.37} & \textbf{27.68} & 27.32 & \underline{16.78} & 17.25 & 17.99 \\
    & AT+MPPD & 42.77 & 24.11 & 22.25 & 21.98 & \underline{10.93} & 12.70 & 11.14 \\
    & AT+SR & 43.10 & 27.04 & 25.65 & 25.46 & \underline{17.22} & 17.70 & 18.55 \\
    & TRADES & 47.76 & 27.68 & 25.85 & 26.08 & 16.61 & \underline{16.59} & 20.39 \\
 \cline{2-9}
    & RTE & \textbf{50.46} & {28.20} & {26.56} & {26.19} & \underline{17.80} & {19.17} & {22.24} \\
    & RTE+SR & 48.78 & 28.67 & {27.48} & \textbf{27.39} & \underline{\textbf{20.15}} & \textbf{20.26} & \textbf{25.04} \\
      \hline
  \end{tabular}
  \caption{Comparison with the state-of-the-art training methods. Square attack is a black-box attack while other attacks are white-box attacks. The worst-case accuracy across all attacks is underlined and used as the robustness evaluation metric.}
  \label{tab1}
\end{table*}

\section{Experiments}

We evaluate our method on four benchmark datasets: SVHN~\cite{netzer2011reading}, CIFAR-10, CIFAR-100, and Tiny-ImageNet\cite{deng2009imagenet}. We use VGG-9~\cite{duan2022temporal} for SVHN, SEWResNet-19~\cite{fang2021deep} for CIFAR-10/100, and SEWResNet-20 for Tiny-ImageNet. Unless otherwise specified, all SNNs are configured with a membrane decay rate $\lambda = 0.5$ and a spiking threshold $V_{\text{th}} = 0.5$. Models are trained on NVIDIA A100 GPUs, and detailed hyperparameters are provided in Appendix B.

We assess adversarial robustness under $l_{\infty}$-bounded perturbations. We adopt standard white-box attacks including FGSM, PGD$_k$, and AutoPGD with both cross-entropy (APGD$^{\text{CE}}$) and DLR (APGD$^{\text{DLR}}$) losses, and use the Square attack~\cite{andriushchenko2020square} for black-box evaluation. Perturbation budgets are set to $\varepsilon = 8/255$ for SVHN and CIFAR, and $\varepsilon = 4/255$ for Tiny-ImageNet. Following~\cite{liu2024enhancing}, we evaluate with both surrogate gradients and rate gradient approximation (RGA)~\cite{bu2023rate}, and report the worst-case accuracy across variants(Detailed results can be found in the Appendix D).
\subsection{Comparison with State-of-the-Art Methods}

We compare our proposed RTE method with two standard baselines, AT and TRADES, as well as several state-of-the-art regularization-based adversarial defense methods for SNNs, including RAT, MPPD, and SR. For fair comparison, the loss function between the model's output and the true label is formulated using the framework of TET~\cite{deng2022temporal}. The training perturbation bounds are set to $\varepsilon = 8/255$ for SVHN, CIFAR-10, and CIFAR-100, and $\varepsilon = 4/255$ for Tiny-ImageNet. Consistent with their original implementations, SR and MPPD are combined with AT (denoted as AT+SR and AT+MPPD). Specific regularization settings are detailed in Appendix C.

As shown in Table~\ref{tab1}, RTE achieves the highest clean accuracy on CIFAR-100 and Tiny-ImageNet, with improvements of 2.53\% and 2.70\%, respectively. On SVHN, its clean accuracy is slightly lower than that of TRADES by 0.41\%. For robustness evaluation, we notice that under weaker white-box attacks such as FGSM and PGD, the measured robust accuracy sometimes exceeds that under the black-box Square attack. This counterintuitive result suggests the presence of gradient obfuscation in SNNs. To ensure reliable evaluation, we adopt APGD—a stronger and adaptive white-box attacker—as the standard method, and report the worst-case accuracy across all attacks as the final robustness measure, with the results underlined in Table~\ref{tab1}.

RTE consistently demonstrates superior robustness on SVHN and Tiny-ImageNet, and performs comparably to the best-performing AT+SR method on CIFAR datasets. When further combined with SR (denoted RTE+SR), our method achieves stronger robustness and surpasses AT+SR on CIFAR-100. Although RTE+SR lags slightly behind AT+SR on CIFAR-10 in worst-case robustness (by 0.90\%), it surpasses AT+SR in clean accuracy by 2.30\%.

\begin{figure}[t]
  \centering
  \includegraphics[width=0.95\columnwidth]{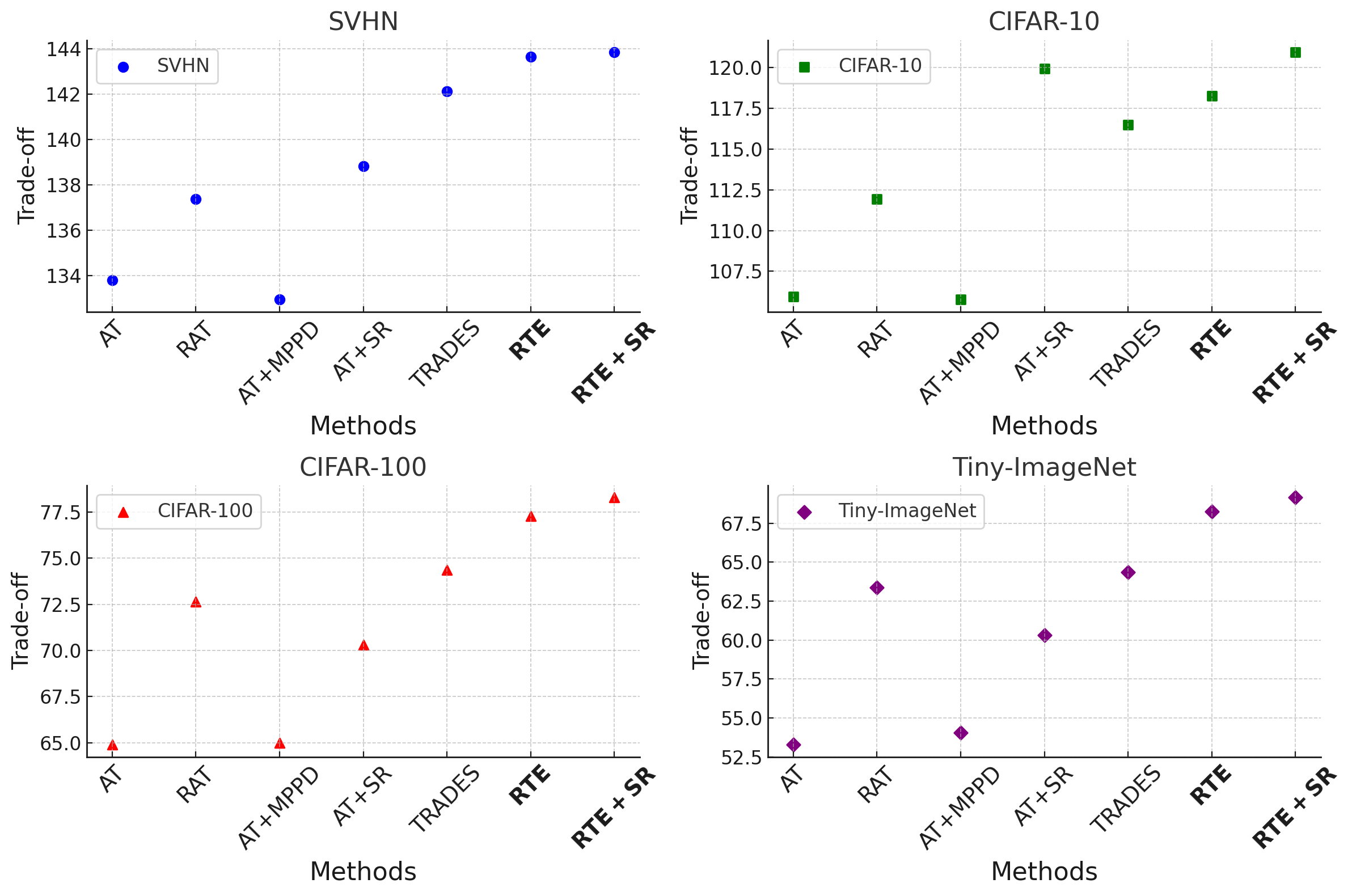}
  \caption{Robustness–accuracy trade-off achieved by different training methods.}
  \label{fig3}
\end{figure}

Importantly, prior methods such as AT+SR often sacrifice clean accuracy for robustness. In contrast, RTE achieves a better trade-off. Following~\cite{wu2024rsc}, we adopt the sum of clean and robust accuracy as the trade-off metric. As illustrated in Figure~\ref{fig3}, RTE+SR achieves the best trade-off on CIFAR-10 and significantly outperforms all baselines on other datasets. Notably, on CIFAR-100 and Tiny-ImageNet, RTE+SR improves the trade-off metric by 3.92 and 4.80 points respectively over the strongest baseline. We attribute these improvements to the enhanced smoothness introduced by SR, which has been shown to facilitate robust ensemble representations~\cite{yang2021trs}.

In terms of training cost, RTE shares similar computational overhead with other methods due to the use of PGD-based adversarial perturbations generation. As detailed in Appendix E, RTE only incurs slightly more cost than AT, making it a practical and scalable defense strategy.

\subsection{Trade-off under Varying Regularization Strength}

\begin{figure}[t]
  \centering
  \includegraphics[width=0.9\columnwidth]{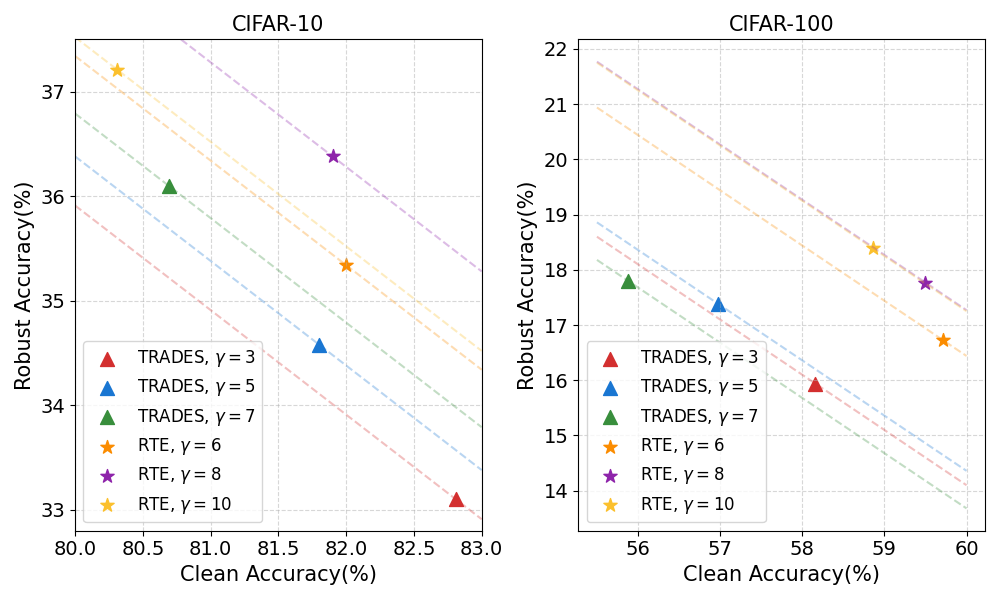}
  \caption{Clean and robust accuracy of RTE (stars) and TRADES (triangles) on CIFAR-10 and CIFAR-100 across different $\gamma$. Dashed lines indicate equal trade-off levels (clean + robust accuracy).}
  \label{fig4}
\end{figure}

To analyze the robustness–accuracy trade-off behavior, we vary the regularization coefficient $\gamma$ for both TRADES and our proposed RTE method. TRADES is selected as a primary comparison baseline due to its wide adoption and explicit trade-off control via $\gamma$. Figure~\ref{fig4} illustrates the clean and robust accuracy on CIFAR-10 and CIFAR-100 under different $\gamma$ values. Each point represents a particular setting of RTE (stars) or TRADES (triangles), and the dashed lines connect points of equal trade-off levels. RTE consistently outperforms TRADES across the trade-off spectrum, achieving more favorable clean–robustness balances, particularly on CIFAR-100. These results demonstrate that RTE provides a consistently better robustness–accuracy trade-off than TRADES across a range of regularization strengths.

\subsection{Effect of Temporal Resolution on RTE}

RTE is explicitly designed to enhance the ensemble effect among temporal sub-networks in SNNs. As the number of timesteps increases, more diverse sub-network predictions contribute to the final output, which can improve adversarial robustness. To examine this effect, we investigate how varying the timestep count $T$ impacts the robustness–accuracy trade-off across four representative training methods: AT, RAT, TRADES, and RTE.

\begin{figure}[htbp]
\centering
\includegraphics[width=0.9\columnwidth]{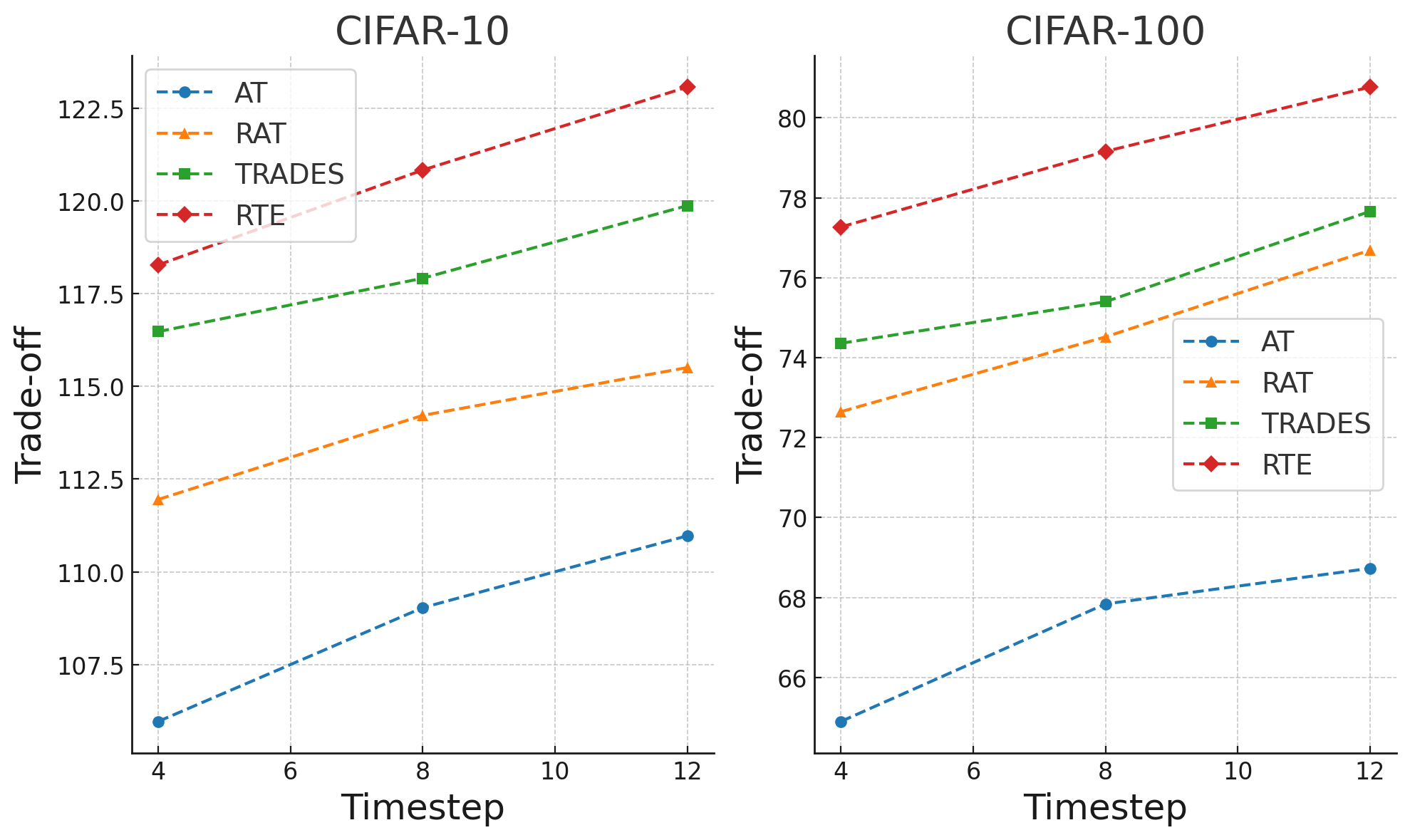}
\caption{Robustness–accuracy trade-off of SNNs trained with different methods under varying timestep counts $T$ on CIFAR-10 and CIFAR-100.}
\label{T}
\end{figure}

As shown in Figure~\ref{T}, increasing $T$ from 4 to 12 consistently improves the trade-off performance for all methods on CIFAR-10 and CIFAR-100. However, RTE consistently achieves the best trade-off across all $T$ values, highlighting its superior ability to leverage temporal diversity. This result suggests that RTE not only benefits from the inherent ensemble nature of SNNs but also scales effectively with increasing temporal resolution.

\subsection{Temporal Robustness Analysis through Transferability Matrices}

To better understand the internal dynamics of adversarial robustness in SNNs' temporal sub-networks, we analyze the transferability matrices of different training methods according to Eq.~\ref{ltm}. 

\begin{figure}[htbp]
\centering
\includegraphics[width=0.9\columnwidth]{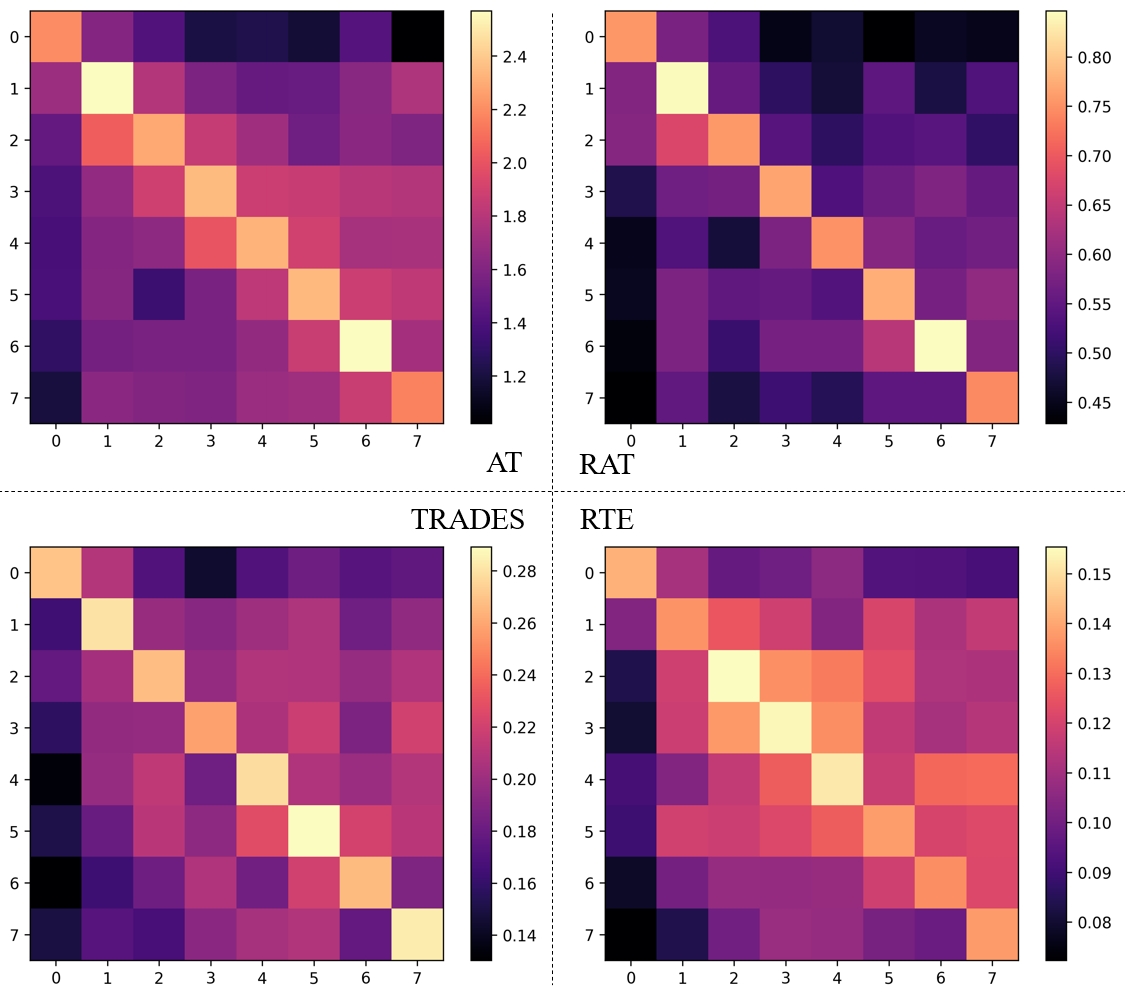}
\caption{Transferability matrices of  AT, RAT, TRADES and RTE. Lower diagonal values indicate better self-robustness of sub-networks, while lower off-diagonal values indicate reduced cross-timestep adversarial transferability.}
\label{fig:transfer}
\end{figure}

Figure~\ref{fig:transfer} presents the resulting matrices for four training strategies: AT, RAT, TRADES and our proposed RTE. The SNNs in the figure are all trained on the CIFAR-10 dataset with a timestep count of 8. The results show that the adversarial transferability between sub-networks shows a decreasing trend as the timestep gap increases, which is consistent with general intuition. Notably, both the diagonal and off-diagonal elements of the transferability matrix for RTE are in the lowest range. These findings highlight the effectiveness of RTE in promoting self-robustness and diversity in temporal sub-networks. By explicitly decoupling vulnerability transferability across timesteps, RTE effectively strengthens the ensemble robustness of SNNs.

In Figure~\ref{fig:transfer} we define $\mathcal{D}$ in Eq.~\ref{ltm} as the KL divergence. We also provide the transferability matrices under $l_2$-norm distance metric in the Appendix F, where the key patterns are consistent with those obtained using KL divergence.


\subsection{Loss Landscape and Temporal Feature Analysis}

To better understand the robustness properties of different training methods, we visualize the output behaviors of SNNs trained with different methods on CIFAR-10 and CIFAR-100 datasets.

\begin{figure}[htbp]
\centering
\includegraphics[width=0.9\columnwidth]{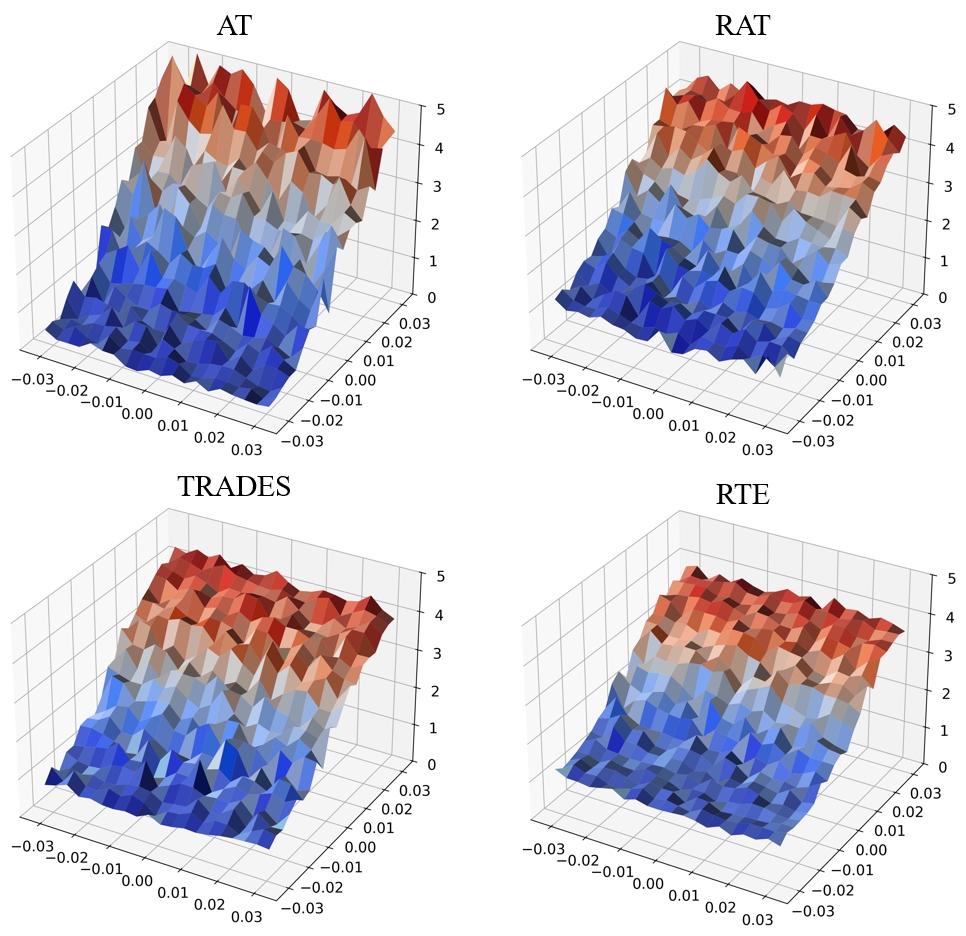}
\caption{Loss surface visualization of SNNs trained with different methods on a CIFAR-100 test sample.}
\label{fig5}
\end{figure}

\paragraph{Loss Surface Visualization.}  
We first visualize the loss landscapes of SNNs trained with different methods on a representative CIFAR-100 test sample. As shown in Figure~\ref{fig5}, the SNN trained with RTE exhibits a significantly smoother and flatter loss surface, along with lower loss values compared to other methods. This suggests that RTE helps stabilize model predictions against input perturbations, indicating stronger adversarial robustness. We also visualize the loss surfaces of sub-networks at different timesteps in Appendix G.

\begin{figure}[htbp]
\centering
\includegraphics[width=0.95\columnwidth]{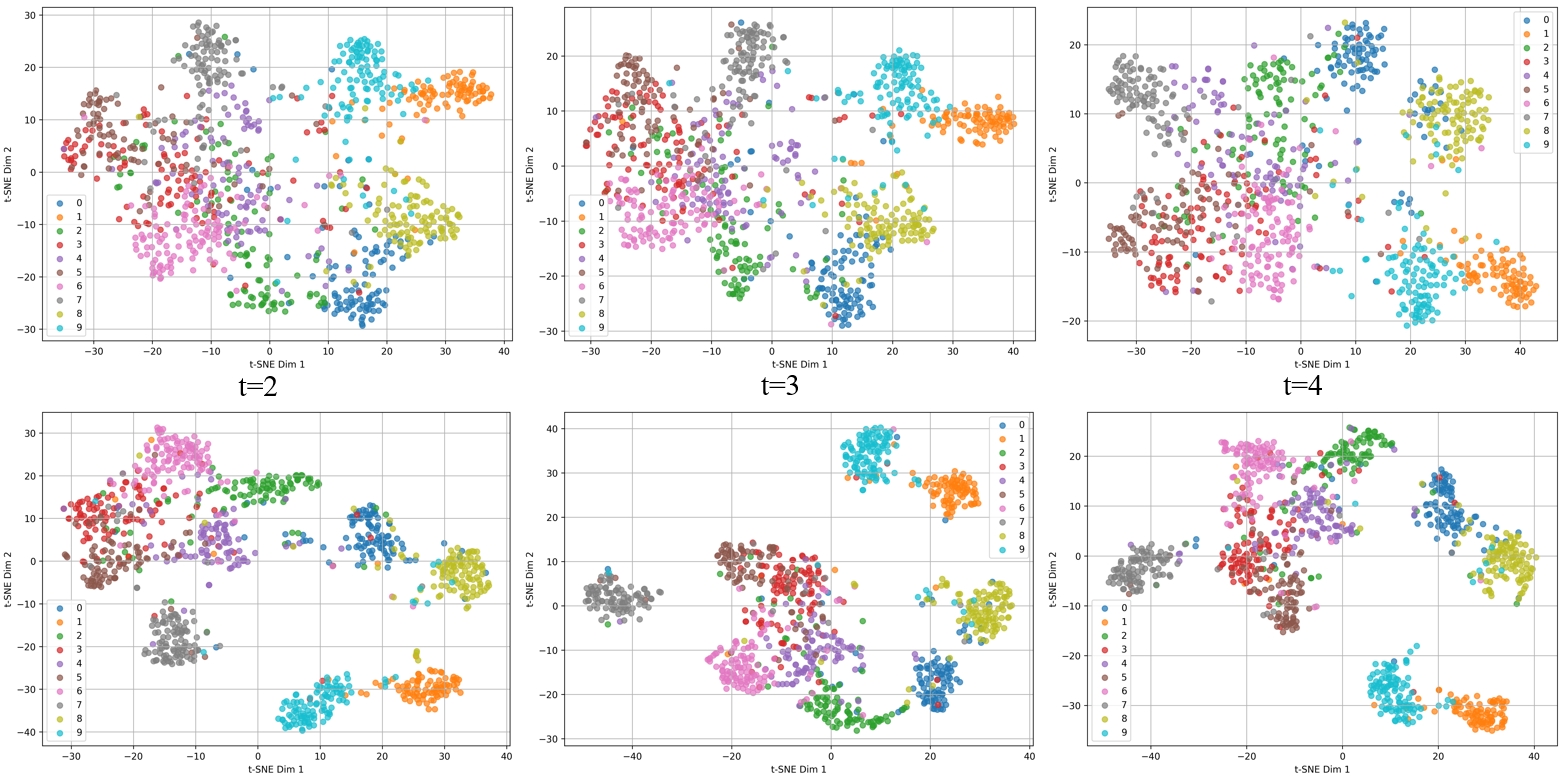}
\caption{t-SNE visualization of final-layer features from sub-networks at different timesteps (2nd–4th) on CIFAR-10. Top: SNN trained with AT; Bottom: SNN trained with RTE.}
\label{fig6}
\end{figure}

\paragraph{Feature Space Visualization.}  
To further evaluate the robustness of temporal sub-networks, we use t-SNE to visualize the final-layer feature distributions at different timesteps. Figure~\ref{fig6} compares the feature embeddings from the 2nd, 3rd, and 4th timesteps of SNNs trained with AT (top row) and RTE (bottom row), using 1,000 CIFAR-10 test samples. The RTE-trained sub-networks consistently exhibit more compact and class-discriminative clusters across timesteps. This demonstrates that RTE improves not only the overall robustness, but also the representation consistency and class separability of individual sub-networks. In Appendix H, we provide additional visualizations of the feature space after temporal ensemble for more methods.

\section{Conclusion}

In this work, we revisit adversarial robustness in SNNs through the lens of temporal self-ensembling. By interpreting SNNs as a sequence of temporally evolving sub-networks, we reveal two intertwined robustness challenges, the insufficient resilience of individual sub-networks and the vulnerability propagation across timesteps. To address these issues, we propose Robust Temporal self-Ensemble (RTE), a novel training framework that simultaneously strengthens each sub-network's robustness and suppresses the temporal alignment of adversarial vulnerabilities.  RTE introduces a principled loss formulation inspired by ensemble learning and employs a stochastic optimization strategy to maintain computational efficiency. Empirical results across four benchmarks demonstrate that RTE consistently achieves state-of-the-art robustness–accuracy trade-offs, while significantly enhancing the diversity and independence among temporal sub-networks. Our analyses further show that RTE reshapes the internal robustness landscape of SNNs by reducing cross-timestep vulnerability transfer, ultimately fostering more reliable and adversarially robust spiking models.


\bibliography{aaai2026}

\begin{thebibliography}{36}
\providecommand{\natexlab}[1]{#1}

\bibitem[{Andriushchenko et~al.(2020)Andriushchenko, Croce, Flammarion, and Hein}]{andriushchenko2020square}
Andriushchenko, M.; Croce, F.; Flammarion, N.; and Hein, M. 2020.
\newblock Square attack: a query-efficient black-box adversarial attack via random search.
\newblock In \emph{European conference on computer vision}, 484--501. Springer.

\bibitem[{Bu et~al.(2023)Bu, Ding, Hao, and Yu}]{bu2023rate}
Bu, T.; Ding, J.; Hao, Z.; and Yu, Z. 2023.
\newblock Rate gradient approximation attack threats deep spiking neural networks.
\newblock In \emph{Proceedings of the IEEE/CVF Conference on Computer Vision and Pattern Recognition}, 7896--7906.

\bibitem[{Croce and Hein(2020)}]{croce2020reliable}
Croce, F.; and Hein, M. 2020.
\newblock Reliable evaluation of adversarial robustness with an ensemble of diverse parameter-free attacks.
\newblock In \emph{International conference on machine learning}, 2206--2216. PMLR.

\bibitem[{Deng et~al.(2009)Deng, Dong, Socher, Li, Li, and Fei-Fei}]{deng2009imagenet}
Deng, J.; Dong, W.; Socher, R.; Li, L.-J.; Li, K.; and Fei-Fei, L. 2009.
\newblock Imagenet: A large-scale hierarchical image database.
\newblock In \emph{2009 IEEE conference on computer vision and pattern recognition}, 248--255. Ieee.

\bibitem[{Deng et~al.(2022)Deng, Li, Zhang, and Gu}]{deng2022temporal}
Deng, S.; Li, Y.; Zhang, S.; and Gu, S. 2022.
\newblock Temporal Efficient Training of Spiking Neural Network via Gradient Re-weighting.
\newblock In \emph{International Conference on Learning Representations}.

\bibitem[{Deng and Mu(2023)}]{deng2023understanding}
Deng, Y.; and Mu, T. 2023.
\newblock Understanding and improving ensemble adversarial defense.
\newblock \emph{Advances in Neural Information Processing Systems}, 36: 58075--58087.

\bibitem[{Ding et~al.(2022)Ding, Bu, Yu, Huang, and Liu}]{ding2022snn}
Ding, J.; Bu, T.; Yu, Z.; Huang, T.; and Liu, J. 2022.
\newblock Snn-rat: Robustness-enhanced spiking neural network through regularized adversarial training.
\newblock \emph{Advances in Neural Information Processing Systems}, 35: 24780--24793.

\bibitem[{Ding et~al.(2024{\natexlab{a}})Ding, Pan, Liu, Yu, and Huang}]{ding2024robust}
Ding, J.; Pan, Z.; Liu, Y.; Yu, Z.; and Huang, T. 2024{\natexlab{a}}.
\newblock Robust Stable Spiking Neural Networks.
\newblock In \emph{Proceedings of the 41st International Conference on Machine Learning}, volume 235 of \emph{Proceedings of Machine Learning Research}, 11016--11029. PMLR.

\bibitem[{Ding et~al.(2024{\natexlab{b}})Ding, Yu, Huang, and Liu}]{ding2024enhancing}
Ding, J.; Yu, Z.; Huang, T.; and Liu, J.~K. 2024{\natexlab{b}}.
\newblock Enhancing the robustness of spiking neural networks with stochastic gating mechanisms.
\newblock In \emph{Proceedings of the AAAI Conference on Artificial Intelligence}, volume~38, 492--502.

\bibitem[{Ding et~al.(2025)Ding, Zuo, Jing, He, and Deng}]{ding2025rethinking}
Ding, Y.; Zuo, L.; Jing, M.; He, P.; and Deng, H. 2025.
\newblock Rethinking spiking neural networks from an ensemble learning perspective.
\newblock \emph{arXiv preprint arXiv:2502.14218}.

\bibitem[{Duan et~al.(2022)Duan, Ding, Chen, Yu, and Huang}]{duan2022temporal}
Duan, C.; Ding, J.; Chen, S.; Yu, Z.; and Huang, T. 2022.
\newblock Temporal effective batch normalization in spiking neural networks.
\newblock \emph{Advances in Neural Information Processing Systems}, 35: 34377--34390.

\bibitem[{Fang et~al.(2021)Fang, Yu, Chen, Huang, Masquelier, and Tian}]{fang2021deep}
Fang, W.; Yu, Z.; Chen, Y.; Huang, T.; Masquelier, T.; and Tian, Y. 2021.
\newblock Deep residual learning in spiking neural networks.
\newblock \emph{Advances in Neural Information Processing Systems}, 34: 21056--21069.

\bibitem[{Goodfellow et~al.(2016)Goodfellow, Bengio, Courville, and Bengio}]{goodfellow2016deep}
Goodfellow, I.; Bengio, Y.; Courville, A.; and Bengio, Y. 2016.
\newblock \emph{Deep learning}, volume~1.
\newblock MIT Press.

\bibitem[{Goodfellow, Shlens, and Szegedy(2015)}]{goodfellow2014explaining}
Goodfellow, I.~J.; Shlens, J.; and Szegedy, C. 2015.
\newblock Explaining and harnessing adversarial examples.
\newblock In \emph{International Conference on Learning Representations}.

\bibitem[{Hao et~al.(2023)Hao, Bu, Shi, Huang, Yu, and Huang}]{hao2023threaten}
Hao, Z.; Bu, T.; Shi, X.; Huang, Z.; Yu, Z.; and Huang, T. 2023.
\newblock Threaten spiking neural networks through combining rate and temporal information.
\newblock In \emph{The Twelfth International Conference on Learning Representations}.

\bibitem[{He et~al.(2016)He, Zhang, Ren, and Sun}]{he2016deep}
He, K.; Zhang, X.; Ren, S.; and Sun, J. 2016.
\newblock Deep residual learning for image recognition.
\newblock In \emph{Proceedings of the IEEE conference on computer vision and pattern recognition}, 770--778.

\bibitem[{Krizhevsky, Hinton et~al.(2009)}]{krizhevsky2009learning}
Krizhevsky, A.; Hinton, G.; et~al. 2009.
\newblock Learning multiple layers of features from tiny images.
\newblock \emph{Technical Report}.

\bibitem[{Li et~al.(2023)Li, Shen, Zhao, Zhang, and Zeng}]{li2023firefly}
Li, J.; Shen, G.; Zhao, D.; Zhang, Q.; and Zeng, Y. 2023.
\newblock Firefly: A high-throughput hardware accelerator for spiking neural networks with efficient dsp and memory optimization.
\newblock \emph{IEEE Transactions on Very Large Scale Integration (VLSI) Systems}, 31(8): 1178--1191.

\bibitem[{Liang et~al.(2021)Liang, Hu, Deng, Wu, Li, Ding, Li, and Xie}]{liang2021exploring}
Liang, L.; Hu, X.; Deng, L.; Wu, Y.; Li, G.; Ding, Y.; Li, P.; and Xie, Y. 2021.
\newblock Exploring adversarial attack in spiking neural networks with spike-compatible gradient.
\newblock \emph{IEEE transactions on neural networks and learning systems}, 34(5): 2569--2583.

\bibitem[{Liu et~al.(2024)Liu, Bu, Ding, Hao, Huang, and Yu}]{liu2024enhancing}
Liu, Y.; Bu, T.; Ding, J.; Hao, Z.; Huang, T.; and Yu, Z. 2024.
\newblock Enhancing Adversarial Robustness in {SNN}s with Sparse Gradients.
\newblock In \emph{Proceedings of the 41st International Conference on Machine Learning}, volume 235 of \emph{Proceedings of Machine Learning Research}, 30738--30754. PMLR.

\bibitem[{Lun et~al.(2025)Lun, Feng, Ni, Liang, Wang, Li, Yu, and Cui}]{lun2025towards}
Lun, L.; Feng, K.; Ni, Q.; Liang, L.; Wang, Y.; Li, Y.; Yu, D.; and Cui, X. 2025.
\newblock Towards Effective and Sparse Adversarial Attack on Spiking Neural Networks via Breaking Invisible Surrogate Gradients.
\newblock In \emph{Proceedings of the Computer Vision and Pattern Recognition Conference}, 3540--3551.

\bibitem[{Maass(1997)}]{maass1997networks}
Maass, W. 1997.
\newblock Networks of spiking neurons: the third generation of neural network models.
\newblock \emph{Neural networks}, 10(9): 1659--1671.

\bibitem[{Madry et~al.(2018)Madry, Makelov, Schmidt, Tsipras, and Vladu}]{madry2017towards}
Madry, A.; Makelov, A.; Schmidt, L.; Tsipras, D.; and Vladu, A. 2018.
\newblock Towards Deep Learning Models Resistant to Adversarial Attacks.
\newblock In \emph{International Conference on Learning Representations}.

\bibitem[{Mukhoty, AlQuabeh, and Gu(2025)}]{mukhoty2025improving}
Mukhoty, B.; AlQuabeh, H.; and Gu, B. 2025.
\newblock Improving Generalization and Robustness in SNNs Through Signed Rate Encoding and Sparse Encoding Attacks.
\newblock In \emph{The Thirteenth International Conference on Learning Representations}.

\bibitem[{Mukhoty et~al.(2024)Mukhoty, AlQuabeh, Masi, Xiong, and Gu}]{mukhotycertified}
Mukhoty, B.; AlQuabeh, H.; Masi, G.~D.; Xiong, H.; and Gu, B. 2024.
\newblock Certified Adversarial Robustness for Rate Encoded Spiking Neural Networks.
\newblock In \emph{The Twelfth International Conference on Learning Representations}.

\bibitem[{Netzer et~al.(2011)Netzer, Wang, Coates, Bissacco, Wu, Ng et~al.}]{netzer2011reading}
Netzer, Y.; Wang, T.; Coates, A.; Bissacco, A.; Wu, B.; Ng, A.~Y.; et~al. 2011.
\newblock Reading digits in natural images with unsupervised feature learning.
\newblock In \emph{NIPS workshop on deep learning and unsupervised feature learning}, volume 2011, 4. Granada.

\bibitem[{Pang et~al.(2019)Pang, Xu, Du, Chen, and Zhu}]{pang2019improving}
Pang, T.; Xu, K.; Du, C.; Chen, N.; and Zhu, J. 2019.
\newblock Improving adversarial robustness via promoting ensemble diversity.
\newblock In \emph{International Conference on Machine Learning}, 4970--4979. PMLR.

\bibitem[{Pei et~al.(2019)Pei, Deng, Song, Zhao, Zhang, Wu, Wang, Zou, Wu, He et~al.}]{pei2019towards}
Pei, J.; Deng, L.; Song, S.; Zhao, M.; Zhang, Y.; Wu, S.; Wang, G.; Zou, Z.; Wu, Z.; He, W.; et~al. 2019.
\newblock Towards artificial general intelligence with hybrid Tianjic chip architecture.
\newblock \emph{Nature}, 572(7767): 106--111.

\bibitem[{Roy, Jaiswal, and Panda(2019)}]{roy2019towards}
Roy, K.; Jaiswal, A.; and Panda, P. 2019.
\newblock Towards spike-based machine intelligence with neuromorphic computing.
\newblock \emph{Nature}, 575(7784): 607--617.

\bibitem[{Shen et~al.(2023)Shen, Zhao, Dong, and Zeng}]{shen2023brain}
Shen, G.; Zhao, D.; Dong, Y.; and Zeng, Y. 2023.
\newblock Brain-inspired neural circuit evolution for spiking neural networks.
\newblock \emph{Proceedings of the National Academy of Sciences}, 120(39): e2218173120.

\bibitem[{Wang et~al.(2025)Wang, Zhao, Du, He, Zhang, and Zeng}]{wang2025random}
Wang, J.; Zhao, D.; Du, C.; He, X.; Zhang, Q.; and Zeng, Y. 2025.
\newblock Random heterogeneous spiking neural network for adversarial defense.
\newblock \emph{iScience}, 28(6).

\bibitem[{Wu et~al.(2024)Wu, Yao, Chou, Qiu, Yang, Xu, and Li}]{wu2024rsc}
Wu, K.; Yao, M.; Chou, Y.; Qiu, X.; Yang, R.; Xu, B.; and Li, G. 2024.
\newblock RSC-SNN: Exploring the Trade-off Between Adversarial Robustness and Accuracy in Spiking Neural Networks via Randomized Smoothing Coding.
\newblock In \emph{Proceedings of the 32nd ACM International Conference on Multimedia}, 2748--2756.

\bibitem[{Xu et~al.(2024)Xu, Ma, Tang, Zheng, and Pan}]{xu2024feel}
Xu, M.; Ma, D.; Tang, H.; Zheng, Q.; and Pan, G. 2024.
\newblock FEEL-SNN: Robust spiking neural networks with frequency encoding and evolutionary leak factor.
\newblock \emph{Advances in Neural Information Processing Systems}, 37: 91930--91950.

\bibitem[{Yang et~al.(2020)Yang, Zhang, Dong, Inkawhich, Gardner, Touchet, Wilkes, Berry, and Li}]{yang2020dverge}
Yang, H.; Zhang, J.; Dong, H.; Inkawhich, N.; Gardner, A.; Touchet, A.; Wilkes, W.; Berry, H.; and Li, H. 2020.
\newblock Dverge: diversifying vulnerabilities for enhanced robust generation of ensembles.
\newblock \emph{Advances in Neural Information Processing Systems}, 33: 5505--5515.

\bibitem[{Yang et~al.(2021)Yang, Li, Xu, Zuo, Chen, Zhou, Rubinstein, Zhang, and Li}]{yang2021trs}
Yang, Z.; Li, L.; Xu, X.; Zuo, S.; Chen, Q.; Zhou, P.; Rubinstein, B.; Zhang, C.; and Li, B. 2021.
\newblock TRS: Transferability Reduced Ensemble via Promoting Gradient Diversity and Model Smoothness.
\newblock In Ranzato, M.; Beygelzimer, A.; Dauphin, Y.; Liang, P.; and Vaughan, J.~W., eds., \emph{Advances in Neural Information Processing Systems}, volume~34, 17642--17655. Curran Associates, Inc.

\bibitem[{Zhao et~al.(2025)Zhao, Shen, Dong, Li, and Zeng}]{zhao2025improving}
Zhao, D.; Shen, G.; Dong, Y.; Li, Y.; and Zeng, Y. 2025.
\newblock Improving stability and performance of spiking neural networks through enhancing temporal consistency.
\newblock \emph{Pattern Recognition}, 159: 111094.

\end{thebibliography}

\end{document}